\documentclass[sigconf]{acmart}

\makeatletter
\renewcommand{\@authorfont}{\normalsize\bfseries}
\renewcommand{\@affiliationfont}{\normalsize\normalfont}
\makeatother

\usepackage{booktabs}
\usepackage{tabularx}
\usepackage{array}
\usepackage{placeins}

\usepackage{tikz}
\usetikzlibrary{arrows.meta,calc,positioning}
\usepackage{xcolor}

\definecolor{domainfill}{HTML}{E7F0F8}
\definecolor{techblue}{HTML}{0000CC}
\definecolor{citered}{HTML}{D00000}
\definecolor{arrowgray}{HTML}{888888}

\graphicspath{{figures/}}
\newcolumntype{Y}{>{\raggedright\arraybackslash}X}
\newcolumntype{L}[1]{>{\raggedright\arraybackslash}p{#1}}
\setcopyright{none}
\renewcommand\footnotetextcopyrightpermission[1]{}
\acmConference[ICAIF '26]{7th ACM International Conference on AI in Finance}{November 14--17, 2026}{Milan, Italy}
\acmYear{2026}
\copyrightyear{2026}
\acmDOI{}
\acmISBN{}

\begin{document}
\title{Auditable Financial Knowledge Graphs: Schema-Driven Construction and Intrinsic Quality Evaluation}


\author{Samarth Sarin}
\affiliation{
  \small
  \institution{BlackRock, Inc.}
  \city{Gurgaon}
  \state{HR}
  \country{India}
}
\email{samarth.sarin@blackrock.com}

\author{Lovepreet Singh}
\affiliation{
  \small
  \institution{BlackRock, Inc.}
  \city{Gurgaon}
  \state{HR}
  \country{India}
}
\email{lovepreet.singh@blackrock.com}

\author{Dhagash Mehta}
\affiliation{
  \small
  \institution{BlackRock, Inc.}
  \city{New York}
  \state{New York}
  \country{USA}
}
\email{dhagash.mehta@blackrock.com}

\renewcommand{\shortauthors}{Sarin Samarth et al.}

\begin{abstract}
Financial documents combine narrative disclosures, accounting metrics, corporate structures, management commentary, and time-dependent facts, making them difficult to represent using conventional chunk-based retrieval-augmented generation systems. This paper presents an auditable, schema-driven methodology for constructing temporal financial knowledge graphs from page-delimited documents parsed using a general purpose document intelligence service. Overlapping document windows undergo phased large language model extraction of structural entities, quantitative observations, and domain-specific relationships. Candidate graph elements are validated against predefined schemas and consolidated through scope-aware exact, normalized, fuzzy, embedding-based, and LLM-assisted entity resolution. Temporal merge constraints, deterministic edge deduplication, conflict-aware property management, and fine-grained provenance preserve issuer identity, reporting-period validity, and traceability to source evidence.

The contributions of this work comprise the proposed methodology for financial knowledge graph construction, the creation of an evidence-grounded temporal knowledge graph from heterogeneous financial documents, and an intrinsic framework for evaluating the quality of the resulting graph. The evaluation framework assesses seven weighted dimensions, two hard validation gates, and structural diagnostics. By separating graph-quality assessment from downstream retrieval and answer generation, the framework supports systematic diagnosis and iterative refinement of the constructed graph. The resulting knowledge graph is subsequently used as the retrieval substrate for GraphRAG-based financial question answering, with FinanceBench serving as a downstream evaluation benchmark. This work establishes knowledge graph construction and quality evaluation as explicit, measurable, and auditable components of financial document intelligence.
\end{abstract}



\maketitle

\section{Introduction}
Financial institutions increasingly use Large Language Models will be referred as LLM and Retrieval-Augmented Generation will be referred as RAG from now on to analyze annual reports, quarterly filings, earnings releases, and regulatory disclosures. These documents contain interconnected numerical, textual, organizational, and temporal information that is often distributed across multiple sections. Conventional RAG retrieves semantically similar text chunks but does not explicitly preserve relationships among companies, financial metrics, reporting periods, risks, and corporate events. Graph Retrieval-Augmented Generation which will be referred as GraphRAG from now on addresses this limitation by retrieving structured entities and relationships, supporting multi-hop reasoning, evidence traceability, and more interpretable financial analysis~\cite{lewis2020retrieval,edge2024local}.

However, the reliability of GraphRAG depends on the quality of its underlying knowledge graph which will be referred as KG from now on. Automatically generated graphs may contain hallucinated entities, unsupported or directionally incorrect relationships, schema violations, duplicate or improperly merged nodes, incomplete coverage, and temporal inconsistencies. Most existing work evaluates GraphRAG primarily through downstream question-answering performance, making it difficult to distinguish graph-construction errors from failures in retrieval, reasoning, or answer generation. This motivates an independent and systematic assessment of intrinsic KG quality.

We present a schema-driven framework for constructing and evaluating temporal financial KGs from heterogeneous financial documents. The construction pipeline combines phased LLM extraction with deterministic validation, scope-aware entity resolution, temporal safeguards, conflict-aware property management, and fine-grained provenance. The accompanying evaluation framework assesses factual grounding, hallucination, schema compliance, semantic consistency, completeness, information density, relationship quality, entity disambiguation, temporal consistency, connectivity, and provenance completeness. The evaluated graph is subsequently used as the retrieval substrate for GraphRAG, with the publicly available set of 150 FinanceBench questions serving as a downstream validation benchmark~\cite{islam2023financebench}. By separating intrinsic graph evaluation from question-answering evaluation, the proposed framework supports the development of more reliable, auditable, and explainable financial document intelligence systems.

\section{Background and Related Work}
\subsection{Financial Documents and Financial Question Answering}
Financial documents, including annual reports, quarterly filings, earnings releases, and regulatory disclosures, are important sources of information for investors, analysts, and financial institutions. They combine structured financial statements with business descriptions, management discussion and analysis (MD\&A), legal disclosures, risk factors, and forward-looking statements, making them substantially more challenging to analyze than conventional text corpora. Recent advances in LLMs have accelerated research in financial question answering (QA) and motivated the development of domain-specific benchmarks. In this study, we use the publicly available subset of 150 FinanceBench questions, each grounded in a source financial document and accompanied by a reference answer and supporting evidence~\cite{islam2023financebench}. Its emphasis on evidence-grounded reasoning makes FinanceBench well suited for the downstream evaluation of retrieval-based financial QA and GraphRAG systems.

\subsection{Retrieval-Augmented Generation and GraphRAG}
RAG combines parametric language models with external retrieval systems to improve factual grounding and reduce hallucinations during text generation~\cite{lewis2020retrieval}. While conventional RAG retrieves semantically similar text chunks using dense vector representations, it often struggles with questions requiring multi-hop reasoning or evidence distributed across multiple document sections. GraphRAG extends this paradigm by organizing documents as KGs, enabling retrieval over entities and relationships rather than isolated passages. Microsoft's GraphRAG demonstrated that graph-based indexing substantially improves corpus-level reasoning and query-focused summarization over large document collections~\cite{edge2024local}. More recent work has further explored graph-guided retrieval for enterprise applications. HybridRAG integrates graph traversal with vector retrieval to improve information extraction from financial documents, demonstrating superior retrieval and answer quality compared to GraphRAG or VectorRAG individually~\cite{sarmah2024hybridrag}. Similarly, recent financial GraphRAG systems have shown that KGs improve explainability, reduce hallucinations, and significantly reduce token consumption on FinanceBench-based evaluations~\cite{barry2025graphrag}. Recent surveys categorize GraphRAG into graph construction, graph-guided retrieval, and graph-enhanced generation, while identifying graph construction quality as one of the major open research challenges~\cite{peng2025graph}.

\subsection{Knowledge Graph Construction for Financial Documents}
KGs have become an important mechanism for organizing heterogeneous and relational information across enterprise applications. Hogan et al.\ identify schema design, entity identity, ontology alignment, provenance, validation, and information extraction as fundamental challenges in KG construction~\cite{hogan2021knowledgegraphs}. Financial documents introduce additional complexities because entities must be resolved across legal names, ticker symbols, reporting entities, fiscal periods, accounting standards, and numerical observations. Consequently, recent financial GraphRAG systems emphasize schema-guided extraction, entity resolution, provenance preservation, temporal reasoning, and graph refinement to improve retrieval reliability~\cite{sarmah2024hybridrag,barry2025graphrag}. Nevertheless, most existing systems evaluate the effectiveness of graph construction indirectly through downstream retrieval or question-answering performance rather than assessing whether the constructed graph itself faithfully represents the source documents.

\subsection{Knowledge Graph Evaluation}

Existing GraphRAG studies primarily assess downstream retrieval and generation using metrics such as answer correctness, faithfulness, context precision and recall, Recall@K, and F1-score, while treating the underlying graph as an intermediate artifact~\cite{peng2025graph}. General KG research considers correctness, completeness, consistency, provenance, and structural validity~\cite{hogan2021knowledgegraphs}; however, these dimensions do not fully capture financial requirements such as numerical and temporal consistency, schema compliance, entity resolution, relationship correctness, and evidence traceability.

Recent research demonstrates the significance of high-quality graph-grounded evidence in financial reasoning, as shown by FinReflectKG~\cite{arun2025finreflectkg}. Moreover, while emerging GraphRAG benchmarks primarily compare retrieval strategies, we aim to create a comprehensive framework for evaluating automatically generated financial KGs by assessing factual grounding, hallucination, schema adherence, provenance completeness, entity resolution, coverage, connectivity, relationship quality, and temporal consistency, thereby enabling systematic graph improvement before their deployment as GraphRAG systems.

\section{Dataset and Problem Formulation}

\subsection{Dataset}
Our experiments are conducted using the FinanceBench benchmark, a publicly available dataset for evaluating financial question-answering systems over real-world corporate disclosures~\cite{islam2023financebench}. FinanceBench comprises 10,231 question–answer–evidence triplets across 40 U.S. public companies and 368 public filings with 143 pages on average per pdf, released from 2015–2023, including 10-K, 10-Q, 8-K, and earnings reports. We evaluate on its open-source 150-case subset with human-annotated gold answers and evidence. Each question is associated with one or more evidence passages, allowing retrieval systems to be evaluated not only on answer correctness but also on evidence grounding and traceability ~\cite{islam2023financebench}.

To construct the KG, each financial document is converted into a page-delimited plain-text representation while preserving document structure like tables and page boundaries as provenance anchors. Formally, a financial document is represented as an ordered sequence of pages:
\begin{equation}
D=\{p_{1},p_{2},\ldots,p_{n}\},
\end{equation}
where each page $p_i$ corresponds to a contiguous section of the original filing or transcript. The documents encompass diverse reporting formats and disclosure styles, requiring the extraction pipeline to generalize across multiple financial document types rather than a single reporting format.

\subsection{Problem Formulation}
\label{sec:problem_formulation}

Given a financial document $D$, the objective is to construct a schema-driven, temporal, and auditable KG

\begin{equation}
G=(V,E,P,\Pi,T,S),
\end{equation}

where $V$ and $E$ denote typed entity nodes and directed relationships, $P$ contains their properties, $\Pi$ stores provenance, $T$ captures temporal context, and $S$ defines entity scope. The graph must conform to the predefined financial schema, preserve issuer-specific identity, prevent incompatible cross-company or cross-period merges, retain distinct observations across reporting periods, and maintain traceability from every graph element to its source evidence.

Formally, the construction process is represented as

\begin{equation}
f:D\rightarrow G,
\end{equation}

where $f$ should produce a graph that faithfully represents the source document while maximizing intrinsic quality across factual grounding, hallucination control, schema compliance, provenance completeness, entity resolution, temporal consistency, coverage, and connectivity.

The resulting graph is subsequently used as the retrieval substrate for GraphRAG-based financial question answering. For a question $q$, the system retrieves a relevant document-scoped subgraph and generates an evidence-grounded answer. Intrinsic graph evaluation is performed independently of downstream question-answering evaluation, allowing graph-construction errors to be distinguished from retrieval, numerical reasoning, and answer-generation failures.


\section{Knowledge Graph Construction Pipeline}
\label{sec:kg_construction}
Financial documents combine narrative disclosures, numerical statements, legal information, corporate events, and time-dependent observations. A financial KG must therefore preserve not only entity and relation semantics, but also numerical context, temporal validity, provenance, and issuer-specific identity. We address these requirements through a schema-driven pipeline that couples LLM-based extraction with deterministic validation, scope-aware entity resolution, temporal compatibility checks, and provenance-preserving graph construction. LLM outputs are never inserted directly: each entity and relation is treated as a candidate that must satisfy structural, semantic, and temporal constraints before it is committed to the graph.

\subsection{Pipeline Overview}
Each source document is converted into page-indexed text using Azure Document Intelligence (model=prebuilt-layout), which also helps in extracting structured tables, while retaining page and document identifiers. Consecutive pages are concatenated into overlapping contextual windows with 3 pages per window and stride 1, enabling facts that cross page boundaries to be observed in more than one extraction context. Each window is processed in parallel for structural entities, quantitative entities, and semantic relations using GPT-5.4~\cite{openai2026gpt54} with temperature=0, no reasoning, no explicit output token limit, generally 3 retries, and no random seed. Entity resolution uses the text-embedding-3-large embedding model with embedding similartiy threshold=0.88 and fuzzy match similarity threshold=0.7. The resulting candidates are validated against the financial schema, filtered for extraction noise, and deduplicated within the window. Accepted candidates then undergo scope-aware entity resolution, endpoint resolution, edge deduplication, temporal reconciliation, and provenance enrichment. After all windows have been processed, document-level coreference resolution consolidates aliases, while orphan linking and consistency checks improve graph connectivity without deleting isolated but source-supported observations. The final output is an auditable, temporal KG that is evaluated independently before being used as the retrieval substrate for GraphRAG.

\begin{figure*}[t]
\centering
\resizebox{0.98\textwidth}{!}{
\begin{tikzpicture}[
    node distance=0.30cm,
    stage/.style={
        rectangle,
        draw,
        rounded corners=1pt,
        align=center,
        minimum width=2.25cm,
        minimum height=0.75cm,
        font=\scriptsize
    },
    arrow/.style={thick,->,>=stealth}
]
\node[stage] (d) {Financial\\documents};
\node[stage, right=of d] (w) {Page-aware\\windows};
\node[stage, right=of w] (s) {Structural\\extraction};
\node[stage, right=of s] (q) {Quantitative\\extraction};
\node[stage, right=of q] (r) {Relation\\extraction};
\node[stage, right=of r] (v) {Schema validation\\and filtering};
\node[stage, right=of v] (e) {Scope-aware and\\temporal resolution};
\node[stage, right=of e] (g) {Coreference, orphan\\linking, and storage};
\node[stage, right=of g] (o) {Auditable\\financial KG};

\draw[arrow] (d)--(w);
\draw[arrow] (w)--(s);
\draw[arrow] (s)--(q);
\draw[arrow] (q)--(r);
\draw[arrow] (r)--(v);
\draw[arrow] (v)--(e);
\draw[arrow] (e)--(g);
\draw[arrow] (g)--(o);
\end{tikzpicture}
}
\caption{Schema-driven financial knowledge graph construction pipeline.}
\label{fig:kg_pipeline}
\Description{Flowchart showing the multi-stage pipeline for constructing financial KGs, starting from financial documents, through page-aware windows, structural and quantitative extraction, relation extraction, schema validation, scope-aware resolution, coreference and orphan linking, to the final auditable financial KG.}
\end{figure*}
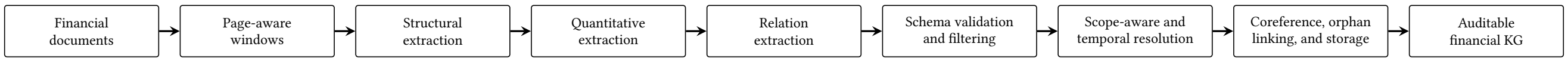

\subsection{Financial Knowledge Graph Schema}
The schema defines the graph objects that may be created and makes construction errors measurable. It contains fourteen entity types divided into structural and quantitative groups, as summarized in Table~\ref{tab:entity_schema}. Structural entities capture the organizational and business context of a document, whereas quantitative entities represent financial and operational observations, instruments, guidance, risks, and temporal references. Each entity type has an associated scope and property contract.

\begin{table*}[t]
\centering
\caption{Financial KG Entity Schema}
\label{tab:entity_schema}
\scriptsize
\setlength{\tabcolsep}{3pt}
\begin{tabular}{@{}p{2.2cm}p{1.55cm}p{1.65cm}p{10.0cm}@{}}
\toprule
\textbf{Entity Type} &
\textbf{Group} &
\textbf{Scope} &
\textbf{Semantics and Representative Properties} \\
\midrule

Company &
Structural &
Global &
Legal reporting or referenced organization; name, ticker, CIK, and industry. \\

Executive &
Structural &
Company &
Senior management or board member; title and role. \\

Product &
Structural &
Company &
Product or product family offered by the company; category and family. \\

Service &
Structural &
Company &
Business service offered by the company; service type. \\

Business Unit &
Structural &
Company &
Organizational division or operating unit; region and business area. \\

Market Segment &
Structural &
Company &
Customer, product, or market segment; segment type. \\

Corporate Event &
Structural &
Company &
Acquisition, restructuring, or related corporate event; event type and date. \\

Location &
Structural &
Global &
City, state, country, or region; geographic type and region. \\

Financial Metric &
Quantitative &
Company &
Financial observation; metric type, value, currency, unit, accounting basis, and period. \\

Operational Metric &
Quantitative &
Company &
Non-financial performance observation; metric type, value, unit, and period. \\

Financial Instrument &
Quantitative &
Company &
Bond, debt, equity, or other security; instrument type and relevant dates. \\

Guidance &
Quantitative &
Company &
Forward-looking quantitative or qualitative outlook; target metric and forecast period. \\

Risk &
Quantitative &
Company &
Business, legal, or regulatory exposure; risk category. \\

Time Reference &
Quantitative &
Company &
Fiscal year, quarter, reporting date, or interval; date and fiscal period. \\

\bottomrule
\end{tabular}
\end{table*}

Node identity is scope-aware. Let $\tau(v)$ denote the entity type, $n(v)$ its normalized name, and $\sigma(v)$ its scope. The canonical lookup key for an entity $v$ is
\begin{equation}
\kappa(v)=\langle \tau(v),\sigma(v),n(v)\rangle.
\label{eq:entity_scope_key}
\end{equation}
For \textit{Company} and \textit{Location}, $\sigma(v)$ is global, allowing the same organization or geography to be reused across documents. For all remaining entity types, $\sigma(v)$ is the normalized filing company. Consequently, identically named concepts such as \textit{Revenue}, \textit{Risk}, or a business segment remain distinct across issuers, while aliases within the same issuer can be consolidated.

Relations are organized into four semantic categories as shown in Table~\ref{tab:relationship_schema}: financial, corporate structure, geography, and legal. Each relation is directional and must connect compatible endpoint types within its category. Edges retain their canonical label, original relation phrase, supporting evidence, confidence, and any temporal or quantitative modifiers.

\begin{table*}[t]
\centering
\caption{Financial KG Schema by Category}
\label{tab:relationship_schema}
\scriptsize
\setlength{\tabcolsep}{3pt}
\begin{tabular}{@{}p{3.0cm}p{4.5cm}p{6.5cm}@{}}
\toprule
\textbf{Category} &
\textbf{Entity Types} &
\textbf{Relation Types} \\
\midrule

Financial &
Company, Financial Metric, Operational Metric, Financial Instrument, Guidance, Time Reference, Business Unit, Market Segment &
\texttt{HAS\_METRIC}, \texttt{DERIVED\_FROM}, \texttt{ISSUED}, \texttt{PAYS\_DIVIDEND}, \texttt{REPURCHASED}, \texttt{PROVIDES\_GUIDANCE\_FOR}, \texttt{SERVES\_SEGMENT} \\

Corporate Structure &
Company, Corporate Event, Executive, Business Unit, Product, Service &
\texttt{ACQUIRER\_IN}, \texttt{TARGET}, \texttt{HAS\_SUBSIDIARY}, \texttt{PARENT\_OF}, \texttt{PRODUCES}, \texttt{OFFERS\_SERVICE} \\

Geography &
Company, Location, Business Unit &
\texttt{HEADQUARTERED\_IN}, \texttt{OPERATES\_IN} \\

Legal &
Company, Risk, Business Unit &
\texttt{SUBJECT\_TO\_LITIGATION}, \texttt{FACES\_INVESTIGATION} \\
\bottomrule
\end{tabular}
\end{table*}

The constrained ontology permits relation-level errors to be distinguished as missing, unsupported, directionally incorrect, or type-invalid edges, thereby providing an explicit basis for the intrinsic KG evaluation described in Section~\ref{sec:evaluation_framework}.

\subsection{Multi-stage Information Extraction}

\paragraph{Structural extraction.}
Each contextual window is first processed using a schema-constrained prompt for the structural types in Table~\ref{tab:entity_schema}. The model must preserve the source surface form and return the entity type, supporting evidence, confidence, and available type-specific properties. Establishing companies, executives, offerings, locations, segments, and corporate events before numerical extraction provides the organizational context required by the subsequent stages.

\paragraph{Quantitative extraction.}
A second pass receives the same window together with a compact representation of its structural entities. It extracts financial and operational metrics, financial instruments, guidance, risks, and temporal references. Financial and operational metrics are accepted only when a usable numerical value is present; units, currency, accounting basis, and reporting period are retained whenever available. Values reported across multiple periods are represented as separate period-specific observations. Financial metrics are also mapped to a controlled metric catalog while their original wording is preserved as evidence.

\paragraph{Relation extraction.}
Relations are extracted in four focused groups---financial, corporate-structure, geographic, and legal---rather than through a single open-ended prompt. Each group receives only the relevant entity types and must reuse previously extracted entity mentions as endpoints. The output includes the canonical relation type, subject--object direction, a concise relation phrase, supporting evidence, confidence, reasoning, and optional date, period, amount, or other modifiers. Restricting extraction by semantic family reduces opportunities for unsupported predicates and type-incompatible relationships.

\paragraph{Validation and window-level consolidation.}
LLM responses are parsed into candidate objects and checked for required fields, valid entity and relation types, confidence ranges, endpoint availability, and type-compatible source--target pairs. Malformed records, unsupported relations, metrics without values, boilerplate filing language, and ambiguous single-word labels are rejected. Within a window, entities are deduplicated using type-aware keys; metric keys additionally preserve metric type, value, unit, and period. Relations are consolidated using their endpoints, canonical type, and modifiers. The highest-confidence candidate is retained, while corroborating evidence is preserved in its provenance.

\subsection{Scope-aware Entity Resolution and Graph Construction}
\label{sec:entity_resolution}

Extracted mentions may contain aliases, abbreviations, or semantically similar names. Entity resolution therefore follows a precision-first cascade. Candidates are initially matched using the scope-aware key $\kappa(v)$, which combines entity type, normalized name, and company scope. Company mentions additionally undergo legal-name and ticker normalization so that abbreviations, tickers, and full legal names resolve to the same global node. Unmatched candidates are compared with same-type, same-scope entities using normalized token similarity, followed by embedding similarity. Substring matching is restricted to suitable types, such as executives and locations, while product candidates require model-number validation and LLM-assisted merge verification. A merge is permitted only when type, scope, and temporal context are compatible; otherwise, a deterministic type-specific identifier is created.

Relationships are inserted only after both textual endpoints have been resolved to graph nodes. Endpoint matching uses exact, case-insensitive, normalized, lexical, and company-aware comparisons, with numerical values and reporting periods used to disambiguate similar metric nodes. For an edge $e=(u,r,v)$ with relation-specific modifiers $m(e)$, the deduplication signature is

\begin{equation}
\delta(e)=\langle u,r,v,m(e)\rangle.
\label{eq:edge_deduplication}
\end{equation}

A new signature creates an edge, whereas a compatible duplicate enriches the existing edge's provenance and updates its confidence only when the incoming observation is stronger. Conflicting observations are preserved as reviewable records rather than discarded. Each accepted edge retains its canonical relation type, original relation phrase, evidence snippet, confidence, reasoning, and relevant temporal or quantitative modifiers.

After document ingestion, type-constrained coreference resolution consolidates remaining aliases and rewrites affected edges. Orphan linking uses page-local evidence to connect isolated nodes where a supported relationship exists. Duplicate, self-referential, and type-invalid edges are removed during refinement, producing a consolidated graph while preserving traceability to the original extraction evidence.

\subsection{Temporal and Provenance-aware Representation}
Temporal compatibility is enforced throughout entity resolution. Two financial metrics may be consolidated only when their metric type, accounting basis, and reporting context are compatible; observations from different fiscal periods must coexist rather than overwrite one another. Time-sensitive values are therefore stored as period-specific properties, whereas stable attributes such as a company's legal identifier or jurisdiction remain unversioned. Temporal metadata is also attached to relations when the underlying statement is valid only for a particular reporting period.

Provenance is maintained at the level of every node, edge, and property observation rather than only at the document level. Each observation records the source document identifier, page, contextual window, supporting evidence snippet, fiscal period, extraction timestamp, confidence score, and run identifier. This representation allows individual graph elements to be traced directly to the passages from which they were derived.

Repeated observations containing the same value are consolidated by unioning their provenance records. When observations disagree, the higher-confidence value remains the primary observation, while the alternative is retained with its own evidence, confidence, and reporting context; no value is silently overwritten. The resulting representation supports source-level auditing, period-aware retrieval, conflict analysis, and direct verification of graph faithfulness within the intrinsic evaluation framework described in Section~\ref{sec:evaluation_framework}.

\section{GraphRAG over Financial Knowledge Graphs}
\label{sec:graphrag}

The constructed KG serves as the retrieval substrate for GraphRAG, replacing independent text chunks with structured, provenance-linked evidence. For each question $q$, retrieval is restricted to the corresponding document graph. Entities, financial concepts, and fiscal periods identified in the question are matched to seed nodes, with \textit{Location}, \textit{Corporate Event}, and temporally compatible \textit{Financial Metric} nodes prioritized for geographic, acquisition-related, and period-specific questions, respectively. The primary \textit{Company} node provides a fallback anchor when no direct match or semantic match is available.

A breadth-first traversal retrieves a document-scoped neighborhood of up to two hops. The resulting context contains entity labels and aliases, canonical relation triples, metric values and units, temporal metadata, page references, confidence scores, and supporting evidence. For computational questions, the system identifies the required metrics and supplies deterministic guidance for ratios, margins, growth rates, percentage changes, and sign handling. The model is instructed to return \textit{insufficient data} when the retrieved graph does not adequately support an answer. If the localized subgraph is insufficient, a recovery pass uses the complete document-scoped graph to improve recall without introducing evidence from unrelated documents.

Downstream evaluation uses the publicly available set of 150 FinanceBench questions~\cite{islam2023financebench}. Generated answers are compared with the corresponding references and graded by an LLM-based evaluator as \textit{Correct} or \textit{Incorrect}. The retrieved graph context, supporting evidence, generated and reference answers, verdict, and token usage are retained for each question, enabling errors to be attributed separately to graph omissions, retrieval, numerical reasoning, or answer generation.

\section{Knowledge Graph Evaluation Framework}
\label{sec:evaluation_framework}
The central premise of this work is that GraphRAG performance depends directly on the quality of the underlying KG. Evaluating only the final question-answering output cannot determine whether an error originated from graph construction, retrieval, numerical reasoning, or answer generation. We therefore evaluate the financial KG independently of the downstream GraphRAG system. The proposed framework combines deterministic graph validation with source-grounded LLM assessment and measures seven complementary dimensions: factual accuracy, semantic consistency, completeness, information density, relationship quality, entity disambiguation, and temporal consistency.

All KG construction, refinement, retrieval, and answer-generation experiments use GPT-5.4~\cite{openai2026gpt54}. GPT-5.5~\cite{openai2026gpt55} is used exclusively as the evaluation model and LLM judge. Separating the construction model from the judge reduces direct self-evaluation bias and allows graph objects to be assessed against the source documents rather than against the model that generated them.

\subsection{Evaluation Protocol}
The evaluation operates on four units: nodes, edges, entity-resolution decisions, and time-bearing property observations. For every evaluated object, the system reconstructs an evidence packet using its stored provenance. The packet contains the original document text, page range, evidence snippet, graph representation, applicable schema definition, reporting period, and confidence score. The evaluator is not provided with the identity of the construction pipeline or its original extraction rationale.

Evaluation proceeds in three stages. First, deterministic rules are applied exhaustively to the graph to verify entity and relation types, mandatory properties, relation domain--range constraints, provenance fields, numeric formats, temporal keys, duplicate identifiers, and graph connectivity. Second, GPT-5.5~\cite{openai2026gpt55} evaluates semantic properties that cannot be reliably determined through rules alone, including factual support, relation direction, contradiction, contextual completeness, and whether two entity mentions refer to the same real-world object. The judge returns a structured decision, error category, confidence score, and concise justification. Unsupported and insufficiently evidenced graph statements are both treated as failures for factual-grounding purposes. Low-confidence or internally inconsistent judgments are submitted for a second adjudication pass.

Third, completeness is measured against an independently generated reference set. GPT-5.5~\cite{openai2026gpt55} extracts schema-valid entities and relations directly from the sampled source pages without access to the constructed graph. The reference objects are subsequently matched to graph objects using normalized names, entity types, endpoint direction, temporal qualifiers, and numerical values. This design prevents the graph from defining its own completeness criterion. Deterministic validation is performed over the entire KG, whereas semantic evaluation is conducted either exhaustively or on a reproducible stratified sample spanning all documents, entity types, and relation types. To ensure balanced evaluation, rare schema categories are retained in their entirety, preventing frequent entity classes from disproportionately influencing the results.

\begin{table*}[!htb]
\centering
\caption{Intrinsic Knowledge Graph Evaluation Dimensions}
\label{tab:kg_evaluation_dimensions}
\scriptsize
\setlength{\tabcolsep}{3pt}
\begin{tabular}{@{}p{2.7cm}p{6.2cm}p{8.1cm}@{}}
\toprule
\textbf{Dimension} &
\textbf{Evaluation Objective} &
\textbf{Operationalization} \\
\midrule
Factual Accuracy
& Determine whether nodes and relations are supported by the source documents.
& Source-grounded GPT-5.5 verification of entity existence, entity type, relation predicate, direction, numerical qualifiers, and supporting evidence. \\

Semantic Consistency
& Detect mutually incompatible graph statements.
& Group comparable claims by entity, predicate or property, scope, and period; identify contradictory relations, incompatible values, and conflicting event dates. \\

Completeness
& Measure whether material document information is represented in the graph.
& Compare graph entities and relations with an independently extracted reference set; additionally measure coverage of important connected entities. \\

Information Density
& Determine whether graph statements retain sufficient context to support reasoning.
& Score each triplet for specificity, standalone interpretability, preservation of material qualifiers, numerical completeness, and absence of boilerplate. \\

Relationship Quality
& Assess whether edges are semantically precise and schema-valid.
& Verify canonical predicate selection, endpoint compatibility, directionality, modifier completeness, and avoidance of generic or fallback relations. \\

Entity Disambiguation
& Measure over-merging and under-merging of entity mentions.
& Evaluate merged alias pairs and high-similarity unmerged pairs; compute pairwise precision, recall, and F1 score. \\

Temporal Consistency
& Verify correct representation of periods and event chronology.
& Check source-aligned period assignment, fiscal-year compatibility, absence of cross-period overwrites, and valid temporal ordering. \\
\bottomrule
\end{tabular}
\end{table*}

\subsection{Intrinsic Quality Metrics}

\paragraph{Factual accuracy and hallucination detection.}
A node is considered grounded only when its existence, semantic type, and material properties are supported by the corresponding source context. An edge is grounded only when the evidence supports both endpoints, the canonical predicate, relation direction, and any associated temporal or quantitative modifiers. Let $P_V$ and $P_E$ denote the proportions of grounded nodes and grounded edges, respectively. Since relations introduce a stronger semantic claim than isolated entity mentions, factual accuracy is defined as
\begin{equation}
FA = 0.4P_V + 0.6P_E.
\label{eq:factual_accuracy}
\end{equation}
This metric directly measures entity and relation hallucination. An entity mentioned elsewhere in the document but unsupported by its recorded provenance is not automatically accepted; the stored evidence must justify the graph observation at the required level of specificity.

\paragraph{Semantic consistency.}
Graph statements are grouped into comparable claim sets using canonical entity identifiers, relation or property type, issuer scope, accounting basis, and reporting period. Deterministic checks normalize units, signs, percentages, currencies, and dates before comparing quantitative observations. GPT-5.5 is then used to identify semantic conflicts, such as mutually exclusive relations or incompatible event descriptions. If $N_C$ is the number of contradictory claim groups and $N_M$ is the number of comparable groups, the consistency score is
\begin{equation}
SC = 1-\frac{N_C}{N_M}.
\label{eq:semantic_consistency}
\end{equation}
Observations from different fiscal periods are not treated as contradictions when their temporal scopes are valid and independently preserved.

\paragraph{Completeness and structural coverage.}
Completeness is evaluated against independently extracted reference entities and relations. Entity recall $R_V$ is the proportion of reference entities represented by at least one valid graph node, while relation recall $R_E$ is the proportion of reference relations represented by a compatible directed edge. Structural coverage $C_G$ measures the proportion of important entities---including companies, metrics, risks, and corporate events---that are connected to the document graph rather than isolated. The combined completeness score is
\begin{equation}
CP = 0.45R_V + 0.45R_E + 0.10C_G.
\label{eq:completeness}
\end{equation}
Entity coverage is reported separately because strong entity recall does not necessarily imply that the relationships required for question answering have been captured.

\paragraph{Information density.}
Each sampled triplet is scored on four applicable criteria: whether it is self-contained, whether it preserves material contextual qualifiers, whether numerical observations retain value, unit, currency, and period, and whether the statement avoids generic or boilerplate language. Each criterion receives a binary score in $[0,1]$, and information density is the mean across all applicable criteria and evaluated triplets. This metric penalizes overly generic statements such as $(\textit{Company},\textit{has},\textit{metric})$ while rewarding graph observations that preserve the financial context necessary for downstream reasoning.

\paragraph{Relationship quality and schema compliance.}
Relationship quality evaluates whether each edge uses an approved canonical predicate, connects schema-compatible node types, preserves the correct subject--object direction, and retains the modifiers required to interpret the relationship. The score is the mean of predicate specificity, endpoint-type validity, direction accuracy, and modifier completeness. Generic or fallback relations are penalized even when their endpoints are valid. Node and edge schema compliance are additionally treated as hard validation gates: an out-of-schema object cannot be considered production-ready merely because it is well connected or textually plausible.

\paragraph{Entity disambiguation.}
Entity resolution is assessed using both positive and negative candidate pairs. Positive candidates are aliases that the construction pipeline merged into one node; hard-negative candidates are lexically or semantically similar mentions that remained separate. GPT-5.5 determines whether each pair denotes the same real-world entity under the applicable company scope and temporal context. Pairwise precision measures erroneous merges, pairwise recall measures missed consolidations, and the entity-disambiguation score is their harmonic mean:
\begin{equation}
ED = \frac{2P_{\mathrm{merge}}R_{\mathrm{merge}}}
{P_{\mathrm{merge}}+R_{\mathrm{merge}}}.
\label{eq:entity_disambiguation}
\end{equation}
This procedure captures both over-merging, such as combining distinct product models, and under-merging, such as retaining a ticker and full legal name as separate company nodes.

\paragraph{Temporal consistency.}
All nodes, edges, and properties containing fiscal years, quarters, dates, or temporal intervals are evaluated against their source evidence. Errors include incorrect period assignment, incompatible cross-year merges, overwritten historical values, invalid event ordering, and contradictory temporal qualifiers. If $N_T$ is the number of evaluated time-bearing observations and $N_T^{err}$ is the number containing a temporal error, then
\begin{equation}
TC = 1-\frac{N_T^{err}}{N_T}.
\label{eq:temporal_consistency}
\end{equation}
\paragraph{Structural diagnostics.}
Graph topology is reported through the number of weakly connected components, largest-component coverage, isolated-node ratio, average degree, articulation points, and bridge edges. These measures are used diagnostically rather than as an independent weighted quality dimension. A densely connected graph can still contain incorrect relationships, while a valid financial graph may contain legitimate document-local components. Connectivity is therefore interpreted jointly with factual accuracy, relationship quality, and completeness rather than treated as a sufficient measure of graph quality.

\subsection{Aggregate Quality Score and Production Criteria}
The seven dimensions are combined into a weighted KG quality score:
\begin{equation}
\begin{split}
KQ ={}& 0.25FA + 0.15SC + 0.15CP + 0.15ID\\
     &+ 0.15RQ + 0.10ED + 0.05TC,
\end{split}
\label{eq:overall_kg_quality}
\end{equation}
where all component scores are normalized to $[0,100]$. Factual accuracy receives the largest weight because an unsupported graph cannot be considered reliable regardless of its coverage or connectivity. Semantic consistency, completeness, information density, and relationship quality receive equal secondary weights, while entity disambiguation and temporal consistency account for the remaining score.

The weighted score is complemented by two hard gates. First, at least 95\% of active nodes and edges must conform to the defined entity and relation schemas. Second, at least 95\% must contain complete document, page, evidence, run, and confidence provenance. A graph failing either gate is not classified as production-ready, even when its aggregate score exceeds the acceptance threshold.

\subsection{GraphRAG Evaluation and Failure Attribution}
Downstream evaluation is performed on 150 FinanceBench questions. For each question, the GraphRAG system scopes the graph to the relevant financial document, selects question-aware seed nodes, retrieves an evidence neighborhood of up to two hops, and invokes the full document-scoped graph when the initial neighborhood is insufficient. GPT-5.4~\cite{openai2026gpt54} generates the final answer using the retrieved graph context and deterministic calculation guidance where required.

GPT-5.5~\cite{openai2026gpt55} evaluates each response against the FinanceBench reference answer and assigns one of two labels: \textit{Correct} or \textit{Incorrect}. The system correctly answers 123 of 150 questions, producing an accuracy of
\begin{equation}
A_{\mathrm{QA}}=\frac{123}{150}=82.00\%.
\label{eq:graphrag_accuracy}
\end{equation}
For each question, the evaluation record retains the selected nodes and edges, source evidence, generated answer, reference answer, judge verdict, and token usage. Unsuccessful cases are classified as graph omissions, unsupported graph facts, retrieval misses, calculation errors, or answer-generation errors. This decomposition prevents downstream model failures from being incorrectly attributed to the graph.

To examine the relationship between intrinsic graph quality and GraphRAG performance, the framework also computes a document-level quality score $KQ_d$ and associates it with the question-answering accuracy for questions grounded in document $d$. Rank correlation between $KQ_d$ and document-level QA accuracy provides a direct empirical test of the hypothesis that improvements in KG quality translate into improved GraphRAG performance.

\subsection{Evaluation Results}

Table~\ref{tab:kg_evaluation_results} reports the acceptance thresholds and the corresponding observed results. Intrinsic KG quality is evaluated separately from downstream GraphRAG accuracy so that graph-construction quality is not conflated with retrieval, numerical reasoning, or answer-generation performance.

\begin{table}[!htb]
\centering
\caption{Knowledge Graph Evaluation Criteria and Experimental Results}
\label{tab:kg_evaluation_results}
\resizebox{\linewidth}{!}{
\begin{tabular}{@{}l l l l@{}}
\toprule
\textbf{Measure} & \textbf{Role} & \textbf{Acceptance Criterion} & \textbf{Observed Result} \\
\midrule
Node and edge schema compliance & Hard gate & $\geq 95.00\%$ & 99\% \\
Provenance completeness & Hard gate & $\geq 95.00\%$ & 99\% \\
Factual Accuracy & 25\% & $\geq 95.00\%$ & 97\% \\
Semantic Consistency & 15\% & $\geq 95.00\%$ & 96\% \\
Completeness & 15\% & $\geq 90.00\%$ & 92\% \\
Entity Coverage & Diagnostic & $\geq 90.00\%$ & \textbf{98\%} \\
Information Density & 15\% & $\geq 90.00\%$ & 93\% \\
Relationship Quality & 15\% & $\geq 90.00\%$ & 96\% \\
Entity Disambiguation & 10\% & $\geq 95.00\%$ & 99\% \\
Temporal Consistency & 5\% & $\geq 95.00\%$ & 95\% \\
Overall KG Quality & Weighted & $\geq 92.75\%$ & 95\% \\
GraphRAG judge-labeled answer accuracy & Downstream & Reported separately & \textbf{82.00\% (123/150)} \\
\bottomrule
\end{tabular}
}
\end{table}

The constructed KG satisfied or exceeded every predefined intrinsic quality threshold. Schema compliance and provenance completeness reached 99\%, indicating that all evaluated graph elements followed the prescribed ontology and retained the required source metadata. The graph achieved 97\% factual accuracy, 96\% semantic consistency, and 96\% relationship quality, while entity disambiguation reached 99\%. Completeness and information density were measured at 92\% and 93\%, respectively, and temporal consistency met its 95\% acceptance threshold. Together, these dimensions produced an overall KG quality score of 95\%.

The entity coverage score of 98\% indicates that the KG represented most of the important entities identified in the source documents. Downstream GraphRAG evaluation correctly answered 123 of the 150 FinanceBench questions, corresponding to a strict accuracy of 82.00\%. The remaining 27 cases were retained for error attribution across graph omissions, retrieval failures, numerical calculation errors, and answer-generation errors.

\section{Conclusion}
\label{sec:conclusion}

This paper introduced an end-to-end framework for constructing, evaluating, and applying KGs over heterogeneous financial documents. The proposed pipeline combines schema-constrained LLM extraction with deterministic validation, scope-aware entity resolution, temporal safeguards, conflict-aware property management, provenance preservation, and graph refinement. Treating extracted entities and relationships as candidates rather than direct graph mutations reduces hallucinated content, inappropriate entity merges, temporal inconsistencies, and loss of source traceability.

A central contribution is the separation of intrinsic KG evaluation from downstream GraphRAG assessment. The proposed framework evaluates factual accuracy, semantic consistency, completeness, information density, relationship quality, entity disambiguation, temporal consistency, schema compliance, and provenance completeness. The resulting KG achieved 98\% entity coverage and supported correct answers for 123 of 150 FinanceBench questions. Overall, the framework establishes KG quality as an explicit and optimizable foundation for reliable, auditable, and explainable financial GraphRAG systems.

\FloatBarrier

\bibliographystyle{ACM-Reference-Format}
\bibliography{references}

\end{document}